\documentclass[runningheads]{llncs}
\usepackage[T1]{fontenc}
\usepackage{times}
\usepackage{soul}
\usepackage{url}
\usepackage[hidelinks]{hyperref}
\usepackage[utf8]{inputenc}
\usepackage[small]{caption}
\usepackage{graphicx}
\usepackage{amsmath}
\usepackage{booktabs}
\usepackage{algorithm}
\usepackage{algorithmic}

\usepackage{amssymb}
\usepackage{dsfont}
\usepackage{multirow}
\usepackage{subfigure}
\usepackage{color}
\usepackage{cleveref}
\usepackage{wrapfig}
\usepackage{verbatim}
\usepackage{bm}
\usepackage{marvosym}

\begin{document}
\title{Incomplete Multi-View Weak-Label Learning with Noisy Features and Imbalanced Labels}
%
\titlerunning{Incomplete Multi-View Weak-Label Learning}
%
\author{Anonymous}
\author{Zhiwei Li\inst{1}\and
Zijian Yang \inst{1} \textsuperscript{\Letter} \and
Lu Sun\inst{1} \and
Mineichi Kudo\inst{2} \and
Keigo Kimura\inst{2}
}   
\authorrunning{Z. Li et al.}
%

%
\institute{}
\institute{ShanghaiTech University, 393 Middle Huaxia Road, Pudong, Shanghai, China \\
\email{\{lizhw, yangzj, sunlu1\}@shanghaitech.edu.cn} \\
\and
Hokkaido University, Kita 8, Nishi 5, Kita-ku, Sapporo, Hokkaido, Japan \\
\email{\{mine,kimura5\}@ist.hokudai.ac.jp}}

\maketitle              
\begin{abstract}
    A variety of modern applications exhibit multi-view multi-label learning, where each sample has multi-view features, and multiple labels are correlated via common views. 
    Current methods usually fail to directly deal with the setting where only a subset of features and labels are observed for each sample, and ignore the presence of noisy views and imbalanced labels in real-world problems.
    In this paper, we propose a novel method to overcome the limitations. It jointly embeds incomplete views and weak labels into a low-dimensional subspace with adaptive weights,
    and facilitates the difference between embedding weight matrices via auto-weighted Hilbert-Schmidt Independence Criterion (HSIC) to reduce the redundancy. Moreover, it adaptively learns view-wise importance for embedding to detect noisy views, and mitigates the label imbalance problem by focal loss. 
    Experimental results on four real-world multi-view multi-label datasets demonstrate the effectiveness of the proposed method.

\end{abstract}

\section{Introduction}
\label{sec:introduction}

In many real-world applications, samples are often represented by several feature subsets, and meanwhile associated with multiple labels \cite{xu2013survey}.
In addition, it is probably that only a subset of features and labels are observed for each sample. Current related methods \cite{zhu2019improved,li2021concise} usually treat multiple view equally and complete the missing data by encouraging low-rankness, which may not hold in practice. 


\begin{figure}[t]
    \begin{center}
        \includegraphics[width = .9\textwidth]{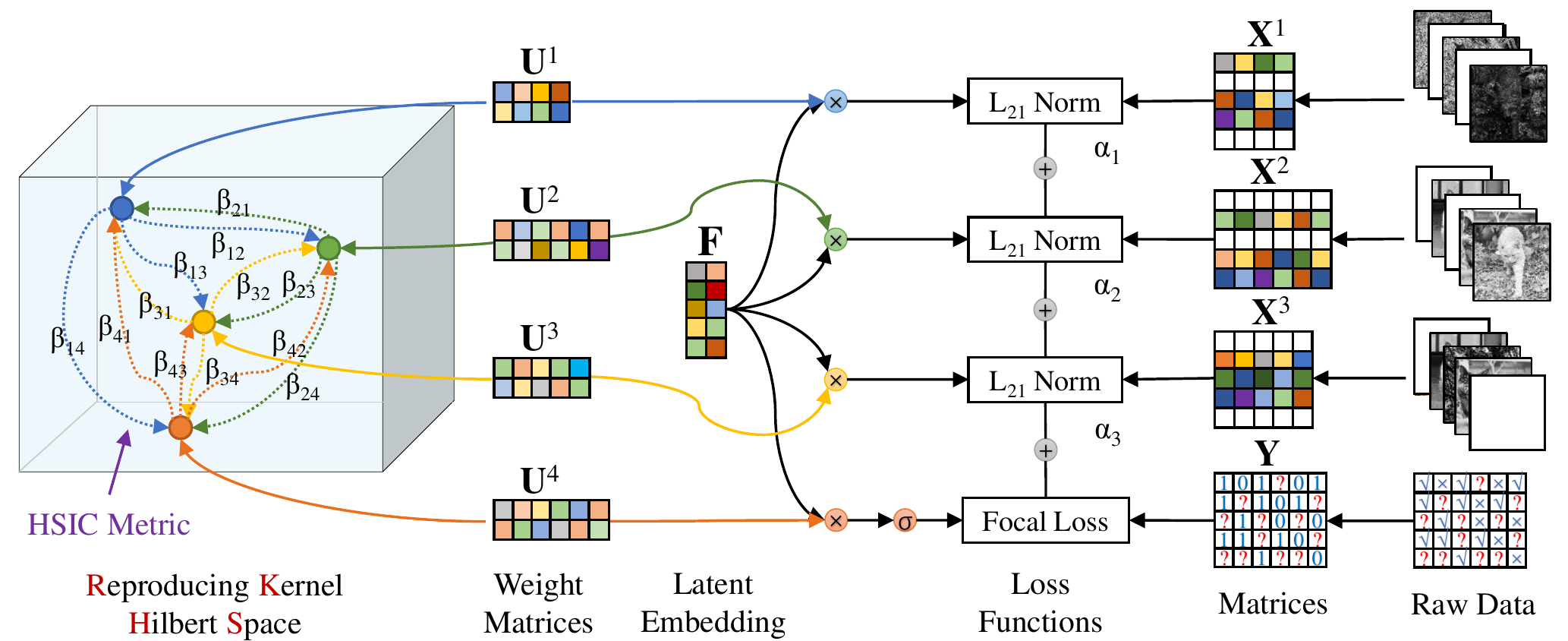}
    \end{center}
    \caption{The framework of NAIL. NAIL first reconstructs incomplete views $\{\mathbf{X}^v\}_{v=1}^m$ and weak labels $\mathbf{Y}$ by a common low-dimensional representation $\mathbf{F}$, i.e., $\mathbf{X}^v \approx \mathbf{FU}^v (\forall v)$ and $\mathbf{Y} \approx \sigma(\mathbf{FU}^{m+1})$, where $\sigma$ denotes the sigmoid function. The reconstruction errors for $\{\mathbf{X}^v\}_{v=1}^m$ and $\mathbf{Y}$ are measured by $L_{2,1}$-norm and focal loss, respectively, and are adaptively weighted by $\{\alpha_v\}_{v=1}^m$. It then projects weight matrices $\{ \mathbf{U}^v\}_{v=1}^{m+1}$ into RKHSs and promotes the differences between weight matrices via $\bm\beta$ auto-weighted HSIC, in order to reduce the redundancy during embedding. Finally, NAIL predicts unobserved labels in $\mathbf{Y}$ based on $\sigma(\mathbf{FU}^{m+1})$.
    }
    \label{fig:workflow}
\end{figure}

To address the challenge, we propose a novel method for i\textbf{N}complete multi-view we\textbf{A}k-label learning with 
no\textbf{I}sy features and imba\textbf{L}anced labels (\textbf{NAIL}). 
NAIL tackles the problem by projecting multiple incomplete views into a common latent subspace using the $L_{2,1}$ norm, adaptively adjusting
view-wise weights to detect noisy views.
It also embeds weak labels into the same subspace, employing Focal Loss to handle label imbalance. 
To remove the redundancy during the embeding, NAIL utilizes the auto-weighted Hilbert-Schmidt Independence Criterion (HSIC) to drive embedding weight matrices to differ from each other in Reproducing Kernel Hilbert Spaces (RKHSs). The workflow of NAIL is illustrated in Fig. 1. 

\section{Methodology}
\label{sec:method}
Let $\mathbf{X}^v = [\mathbf{x}^v_1, \mathbf{x}^v_2, \dots, \mathbf{x}^v_n]^T \in \mathbb{R}^{n \times d_v}$
denote the feature matrix in the $v$-th view, and $\mathbf{Y} = [\mathbf{y}_1, \mathbf{y}_2, \dots, \mathbf{y}_n]^T \in \{0, 1\}^{n \times l}$ denote the label matrix, 
where $y_{ij} = 1$ means that the $j$-th label is assigned to the $i$-th instance and $y_{ij} = 0$ otherwise.
We introduce $\mathbf{O}^v_{\mathbf{X}} \in \mathbb{R}^{n \times d_v}$ and $\mathbf{O}_{\mathbf{Y}} \in \mathbb{R}^{n \times l}$ 
to denote indices of the entries in $\mathbf{X}^v$ and $\mathbf{Y}$, respectively,
such that $(\mathbf{O}^v_{\mathbf{X}})_{ij} = 1$ or $(\mathbf{O}_{\mathbf{Y}})_{ij} = 1$ if the $(i, j)$-th entry is observed in $\mathbf{X}^v$ or $\mathbf{Y}$, 
and $(\mathbf{O}^v_{\mathbf{X}})_{ij} = 0$ or $(\mathbf{O}_{\mathbf{Y}})_{ij} = 0$ otherwise.
The goal of NAIL is to predict unobserved labels in presence of both noisy views and imbalanced labels.

\subsection{Auto-Weighted Incomplete Multi-View Embedding}
Given a multi-view dataset, 
we seek to find a shared latent subspace $\mathbf{F} \in \mathbb{R}^{n \times k}$ ($k < d_v$, $\forall v$) by integrating complementary information from different views \cite{gao2015multi}, which can be formulated as
 $  
        \min_{\mathbf{F}, \{\mathbf{U}^v\} \geq 0}  \sum^m_{v=1}  \ ||\mathbf{X}^v - \mathbf{F} \mathbf{U}^v||^2_F
        $,
where $||\cdot||_F$ represents the Frobenius norm and $\mathbf{U}^v \in \mathbb{R}^{k \times d_v}$ is the weight matrix of the $v$-th view.
It embeds multiple views into an identical subspace by treating each view equally, 
 deviating from the true latent subspace when multiple views have different importance during embedding.
Furthermore, the existence of missing entries poses another challenge.
To address the problems, we propose the auto-weighted incomplete multi-view embedding:
\begin{equation}
    \label{eq:mv_nmf_mask}
        \min_{\substack{\bm\alpha, \mathbf{F}, \{\mathbf{U}^v\} \geq 0, \\ \sum \alpha_v = 1}} \
         \sum^m_{v=1} \mathbf{\alpha}^s_v ||\mathbf{O}^v_{\mathbf{X}} \odot (\mathbf{X}^v - \mathbf{F} \mathbf{U}^v)||_{2,1} 
\end{equation}
where $\odot$ is the Hadamard product, and $||\mathbf{A}||_{2,1} = \sum^n_{i=1} ||\mathbf{a}_{i:}||_2$ represents the $L_{2,1}$ norm, which is insensitive to outlier samples by decreasing the contribution of the outlier to the reconstruction error.
In \eqref{eq:mv_nmf_mask}$, \mathbf{\alpha}_v$ is introduced to weight the embedding importance of the $v$-th view $(\bm\alpha = [\mathbf{\alpha}_1, \mathbf{\alpha}_2, \dots, \mathbf{\alpha}_m])$, and $s$ is a constant, which is fixed as 0.5 in experiments.
According to \eqref{eq:mv_nmf_mask}, $\mathbf{X}^v$ is mapped to a common latent representation $\mathbf{F} \in \mathbb{R}^{n \times k}$ with view-specific adaptive weight $\mathbf{\alpha}_v$.
For the $v$-th view, the more importance  contributed  to embedding $\mathbf{F}$, the higher weight of $\alpha_v$, and vice versa.

\subsection{Imbalanced Weak-Label Embedding}
Cross Entropy (CE) \cite{de2005tutorial} is often used to measure the classification loss between the ground truth and predictions.
However, possible label imbalance, i.e., a large difference between the proportions of positive and negative labels, can lead to a drop in prediction accuracy. 
Here we adopt Focal Loss (FL) \cite{lin2017focal} to mitigate this problem.
For the $j$-th label in the $i$-th sample, focal loss $\text{FL}(y_{ij}, p_{ij})$ is computed based on the ground truth $y_{ij}$ and the predicted label probability $p_{ij}$, i.e.,
$\text{FL}(y_{ij}, p_{ij}) = -a_{ij} (1 - q_{ij})^{\gamma} \log(q_{ij})$,
where $\gamma$ is a constant, and $a_{ij}$ takes a value $a \in [0, 1]$ if $y_{ij} = 1$ and $a_{ij} = 1 - a$ otherwise. 
In experiments, we fix $\gamma = 2$ and $a=0.5$. 
In focal loss, 
$q_{ij}=p_{ij}$ if $y_{ij} = 1$, and $q_{ij}=1 - p_{ij}$ otherwise.
Predicted probability $p_{ij}$ is calculated by
    $p_{ij} = \sigma(\mathbf{{f}}_{i:}^T \mathbf{{u}}^{m+1}_{:j})$,
where $\sigma(\cdot)$ is the sigmoid function, $\mathbf{f}_{i:}$ is the $i$-th row of the latent embedding $\mathbf{F}$ in \eqref{eq:mv_nmf_mask}, and $\mathbf{u}_{:j}^{m+1}$ is the $j$-th column of the weight matrix $\mathbf{U}^{m+1}$ for label embedding.
Therefore, imbalanced weak-label embedding can be modeled as follows:
\begin{equation}
    \label{eq:mask_sum_y}
        \min_{\mathbf{F}, \mathbf{U}^{m+1} \geq 0} \  \sum_{(i,j)\in \mathbf{O}_{Y}} \text{FL}(y_{ij}, \sigma(\mathbf{f}^T_{i:}\mathbf{u}^{m+1}_{:j})). 
\end{equation}
Thus, the label imbalance problem is alleviated by applying focal loss on the observed labels, 
which helps the model to focus on learning hard misclassified samples.

\subsection{Correlation Modeling by Auto-Weighted HSIC}
Next, we adopt the Hilbert-Schmidt Independence Criterion (HSIC) \cite{gretton2005measuring} to model the nonlinear correlations among weight matrices $\{\mathbf{U}^v\}^{m+1}_{v=1}$ in an adaptive manner.
Specifically, HSIC estimates the dependency between $\mathbf{U}^v$ and $\mathbf{U}^{v'}$ ($v' \neq v$) in the Reproducing Kernel Hilbert Spaces (RKHSs), 
   i.e., $ \rm{HSIC}(\mathbf{U}^v, \mathbf{U}^{v'}) = 
   (n-1)^{-2} \rm{tr}(\mathbf{K}^v \mathbf{H} \mathbf{K}^{v'} \mathbf{H})$,
where $\mathbf{K}^v \in \mathbb{R}^{n \times n}$ is the Gram matrix that measures the similarity between row vectors of $\mathbf{U}^v$.
$\mathbf{H} = \mathbf{I} - \frac{1}{n} \mathds{1} \mathds{1}^T$ is the centering matrix,
where $\mathbf{I} \in \mathbb{R}^{n \times n}$ is an identity matrix, and $\mathds{1} \in \mathbb{R}^n$ is an all-one vector.
It is guaranteed that the lower the value of HSIC, the lower the dependence between $\mathbf{U}^v$ and $\mathbf{U}^{v'}$.
Thus, to reduce the redundancy among $\mathbf{U}^v$s during embedding, we can minimize the HSIC between each pair of weight matrices.
However, noisy views make directly minimizing the HSIC too restrictive in practice.
To address the problem, we propose to minimize auto-weighted HSIC, i.e., 
\begin{equation}
    \label{eq:hsic}
        \min_{\substack{\bm{\beta}, \{\mathbf{U}^v\} \geq 0 \\ ||\bm{\beta}_v||_2 = 1}} \  \sum^{m+1}_{v=1} \sum_{v'\neq v}
        \mathbf{\beta}_{vv'} \rm{HSIC}(\mathbf{U}^v, \mathbf{U}^{v'}) 
\end{equation}
where $\mathbf{\beta}_{vv'} \geq 0$ measures the importance of the correlation between 
$\mathbf{U}^v$ and $\mathbf{U}^{v'}$ and $\bm\beta_v = [\mathbf{\beta}_{v1}, \mathbf{\beta}_{v2}, \dots, \mathbf{\beta}_{v(m+1)}]$.
Once the $v$-th view is indeed noisy, a relatively larger value will be assigned to $\beta_{v}$,
leading to the decorrelation between $\mathbf{U}^v$ and $\mathbf{U}^{v'} (\forall v' \neq v)$ by imposing a stronger degree of penalty on HSIC.
Therefore, multiple views and labels are correlated in a non-linear and adaptive way.

\subsection{The Proposed NAIL Method}
By incorporating \eqref{eq:mv_nmf_mask}, \eqref{eq:mask_sum_y} and \eqref{eq:hsic}, we now have the optimization problem of NAIL:
    \begin{align}
        \label{eq:obj}
            & \min_{\substack{\bm{\alpha}, \bm{\beta}, \\ \mathbf{F}, \{\mathbf{U}^v\}} }
             \sum^m_{v=1} \mathbf{\alpha}^s_v ||\mathbf{O}^v_{\mathbf{X}} \odot (\mathbf{X}^v - \mathbf{F} \mathbf{U}^v)||_{2,1} +  \lambda \sum_{(i,j)\in \mathbf{O}_{Y}} \text{FL}(y_{ij}, \sigma(\mathbf{f}^T_{i:}\mathbf{u}_{:j})) \\
            & + \mu \sum^{m+1}_{v=1} \sum_{v'\neq v} \mathbf{\beta}_{vv'} \rm{HSIC}(\mathbf{U}^v, \mathbf{U}^{v'}), \ \ 
            s.t. \  \sum \mathbf{\alpha}_v = 1, ||\bm\beta_v||_2 = 1,  \bm{\alpha}, \bm{\beta}, \mathbf{F}, \{\mathbf{U}^v\} \geq 0, \nonumber
    \end{align}
where $\lambda$ and $\mu$ are nonnegative hyperparameters.
It is worth noting that $\alpha_v$ weights the reconstruction error between $\mathbf{X}^v$ and $\mathbf{FU}^v$, while $\bm\beta_v$ weights the correlation between $\mathbf{U}^v$ and $\mathbf{U}^{v'}$ ($\forall v' \neq v$). In other words, once $\mathbf{X}^v$ is noisy, $\alpha_v$ will be assigned to a small value as it cannot be recovered well by $\mathbf{FU}^v$, while $\bm\beta_v$ will take a large value in order to decorrelate $\mathbf{U}^v$ with $\mathbf{U}^{v'}$ ($\forall v' \neq v$). In this way, NAIL adaptively embeds incomplete views and weak labels into a common latent subspace, and non-linearly decorrelates weight matrices with adaptively weights, enabling to complete missing labels in presence of both noisy views and imbalanced labels. Once \eqref{eq:obj} is solved, the prediction for missing labels is made by thresholding $\sigma(\mathbf{f}^T_{i:}\mathbf{u}_{:j})$ with a threshold of 0.5.

\section{Experiments}
\label{sec:experiments}

\subsection{Experimental Settings}

We conduct experiments on four benchmark multi-view multi-label datasets:
Corel5k\footnote{\url{http://lear.inrialpes.fr/people/guillaumin/data.php}\label{guillaumin}}, Pascal07\footref{guillaumin}, Yeast dataset\footnote{\url{http://vlado.fmf.uni-lj.si/pub/networks/data/}}
and Emotions\footnote{\url{http://www.uco.es/kdis/mllresources}}.
The proposed NAIL\footnote{The code and supplement: \url{https://github.com/mtics/NAIL}} is compared with four state-of-the-art methods:
lrMMC \cite{liu2013multi}, McWL \cite{tan2018multi}, iMVWL \cite{tan2018incomplete} and NAIM$^3$L \cite{li2021concise}.
lrMMC and McWL are adopted by filling missing features with zero, and 
iMVWL and NAIM$^3$L are originally designed for incomplete multi-view weak-label learning.
NAIL uses the Gaussian kernel in HSIC, and NAIL-L is its variant with the linear kernel.

We tune the hyperparameters of lrMMC, NAIL-L and NAIL on all datasets, and tune the hyperparameters of McWL, iMVWL and NAIM$^3$L on the Yeast and Emotions datasets by grid search to produce the best possible results.
On the two image datasets, hyperparameters of McWL, iMVWL and NAIM$^3$L are selected as recommended in the original papers.
We select the values of hyperparameters $\lambda$ and $\mu$ from $\{10^i|i=-3, \dots, 3\}$, 
and the ratio $r_k$ of $\frac{k}{d}$ from \{0.2, 0.5, 0.8\} for NAIL and NAIL-L.
We set $s=a=0.5$  and $\gamma=2$ in experiments.
We randomly sample 2000 samples of each image dataset, and use all samples from the Yeast and Emotions datasets in the experiment.
We randomly remove $r\%$ samples from each feature view by ensuring that each sample appears in at least one feature view, 
and randomly remove $s\%$ positive and negative samples for each label. 
We randomly select 70\% of the datasets as the training set and use the rest as the validation set,
and repeat this procedure by ten times and report the average values and the standard deviations.
The prediction performance is evaluated by two metrics: Hamming Score (HS) \cite{zhang2013review} and Average Precision (AP) \cite{bucak2011multi}.
In this work, our goal is to complete the missing labels in the training set.

\subsection{Experimental Results}

\begin{table}[!t]
    \centering
    \caption{Experimental results on four real-world datasets at $r\% =50\%$ and $s\% =50\%$. The best results are highlighted in boldface, and the second best results are underlined.}
    \resizebox*{1\textwidth}{!}{
        \begin{tabular}{cccccccccccccc}
            \hline
            \multicolumn{2}{c}{\multirow{2}{*}{}} & \multicolumn{2}{c}{lrMMC} & \multicolumn{2}{c}{McWL} & \multicolumn{2}{c}{iMVWL} & \multicolumn{2}{c}{NAIM$^3$L} & \multicolumn{2}{c}{\textbf{NAIL-L}} & \multicolumn{2}{c}{\textbf{NAIL}} \\ \cline{3-14} 
            \multicolumn{2}{c}{}                  & Mean        & STD         & Mean        & STD        & Mean             & STD    & Mean             & STD     & Mean              & STD     & \textbf{Mean}     & STD     \\ \hline
            \multirow{2}{*}{Emotions}    & HS     & 0.5057      & 0.0125      & 0.6303      & 0.0031     & 0.6281           & 0.0082 & 0.6911           & 0.0068  & \underline{0.6920}            & 0.0307  & \textbf{0.7135}   & 0.0104  \\
                                         & AP     & 0.5293      & 0.0140      & 0.6102      & 0.0111     & 0.6006           & 0.0029 & 0.6783           & 0.0149  & \underline{0.6923}            & 0.0291  & \textbf{0.7017}   & 0.0099  \\ \hline
            \multirow{2}{*}{Yeast}       & HS     & 0.7275      & 0.0002      & 0.7420      & 0.0020     & 0.7337           & 0.0113 & 0.7089           & 0.0003  & \textbf{0.7522}   & 0.0049  & \underline{0.7462}            & 0.0081  \\
                                         & AP     & 0.6503      & 0.0000      & 0.6936      & 0.0040     & 0.7219           & 0.0037 & 0.6665           & 0.0113  & \textbf{0.7267}   & 0.0102  & \underline{0.7235 }           & 0.0187  \\ \hline
            \multirow{2}{*}{Corel5k}     & HS     & 0.9084      & 0.0089      & 0.9070      & 0.0001     & 0.9581           & 0.0090 & 0.9575           & 0.0174  & \underline{0.9792}            & 0.0064  & \textbf{0.9800}   & 0.0058  \\
                                         & AP     & 0.1897      & 0.0021      & 0.1527      & 0.0052     & 0.2643           & 0.0005 & \textbf{0.5212}  & 0.0142  & \underline{0.3594}            & 0.0834  & 0.3436            & 0.0028  \\ \hline
            \multirow{2}{*}{Pascal07}    & HS     & 0.9194      & 0.0009      & 0.8132      & 0.0004     & 0.8690           & 0.0144 & 0.9211           & 0.0071  & \underline{0.9450}            & 0.0131  & \textbf{0.9480}   & 0.0096  \\
                                         & AP     & 0.3998      & 0.0013      & 0.3438      & 0.0032     & 0.4364           & 0.0169 & 0.4494           & 0.0076  & \textbf{0.4892}   & 0.0138  & \underline{0.4828}            & 0.0188  \\  \hline
        \end{tabular}
    }
    \label{table:main_results_0.5}
\end{table}

\subsubsection{Evaluation of Comparing Methods}
Table \ref{table:main_results_0.5} shows the experimental results of all comparing methods on four real-world datasets at $r\% = 50\%$ and $s\% = 50\%$.
From Table \ref{table:main_results_0.5}, we can see that NAIL and NAIL-L outperform comparing methods in most of the cases.
The performance superiority probably comes from their ability on handling noisy views and imbalanced labels, 
and decorrelating weight matrices for redundancy removal in an adaptive way.
The incompleteness of multi-view data causes the performance degradation of lrMMC and McWL. iMVWL and NAIM$^3$L outperform lrMMC and McML in most cases, but perform worse than NAIL and NAIL-L.
There are two possible reasons: one is that iMVWL assumes that the label matrix is low-rank,
and the other is that both iMVWL and NAIM$^3$L treat multiple views equally. 
In contrast, NAIL and NAIL-L measure the importance of each view by adaptively choosing appropriate values of $\bm{\alpha}$ and $\bm{\beta}$.
In summary, it shows that once a low-dimensional space indeed contains nonlinear transformations about features and labels, NAIL enables to save their structural properties and uses the HSIC to capture correlations between them.

\begin{wrapfigure}[10]{r}{0.53\linewidth}
    \centering
    \includegraphics[width = 1\linewidth]{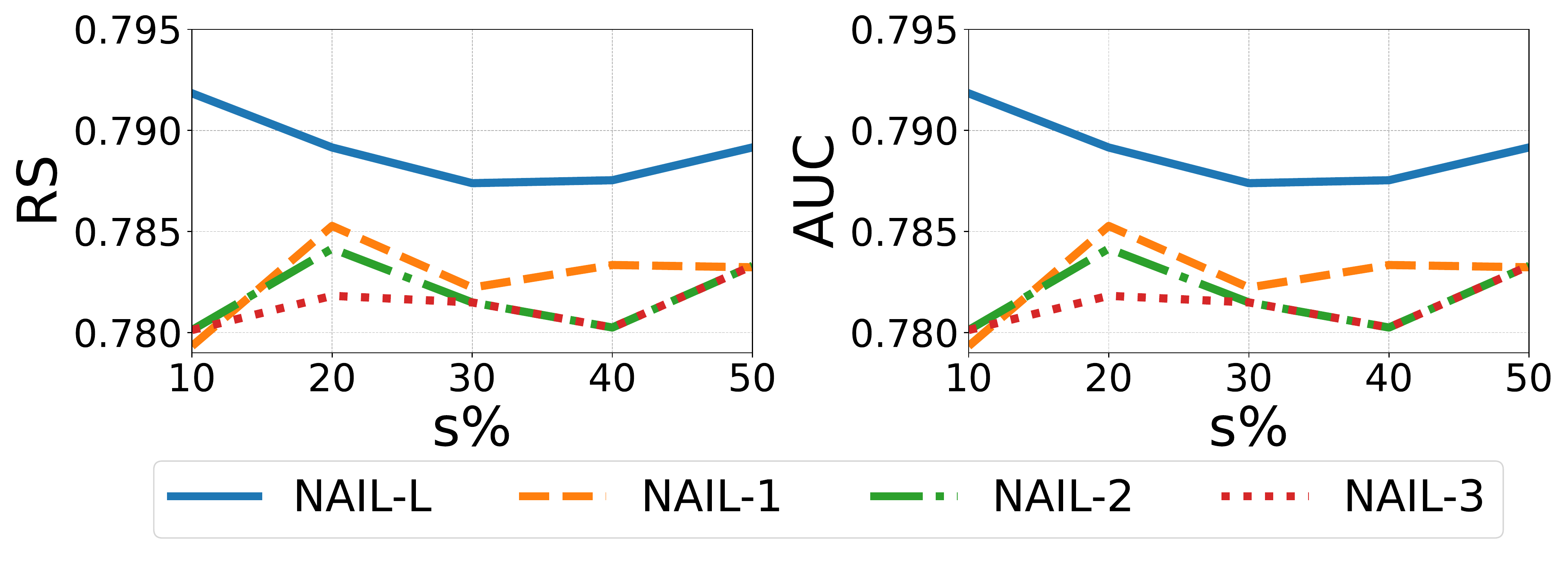}
    \caption{Ablation study of NAIL on the Corel5k dataset at $r\% = 50\%$ by varying $s\%$ from $10\%$ to $50\%$ by step $10\%$.}
    \label{fig:variants_change_label}
\end{wrapfigure}

\subsubsection{Ablation Study}
\label{sec:ablation}
To investigate the effects of NAIL-L's components, we introduce three variants of NAIL-L, 
namely NAIL-1, NAIL-2 and NAIL-3.
NAIL-1 uses Frobenius norm to measure the reconstruction error of features and labels, instead of $L_{2,1}$ norm and focal loss.
NAIL-2 ignores the decorrelation between weight matrices during embedding by simply removing auto-weighted HSIC.
NAIL-3 treats multiple views equally in both reconstruction and decorrelation, by omitting $\bm\alpha$ and $\bm\beta$. 
Fig. \ref{fig:variants_change_label} shows the ablation study of NAIL-L on the Corel5k dataset at $r\% = 50\%$ by varying values of $s\%$.
Among the variants, NAIL-3 performs the worst as it fails to detect noisy views. NAIL-1 and NAIL-2 perform worse than NAIL-L, probably because the simple Frobenius norm based loss in NAIL-1 is sensitive to sample outliers and imbalanced labels, and the removal of HSIC in NAIL-2 is harmful for generalization.
In contrast, NAIL-L has the best performance in RS and AUC on all datasets, indicating the effectiveness and necessity of its components.

\section{Conclusion}
\label{sec:conclusion}

In this paper, we propose a novel method called NAIL to deal with incomplete multi-view weak-label data.
NAIL jointly embeds incomplete views and weak labels into a shared subspace with adaptive weights,
and facilitates the difference between the embedding weight matrices via auto-weighted HSIC.
Moreover, to deal with noisy views and imbalanced labels, adaptive $L_{2,1}$ norm and focal loss are used to calculate the reconstruction errors for features and labels, respectively. 
Empirical evidence verifies that NAIL is flexible enough to handle various real-world problems. 

%
%
%
\bibliographystyle{splncs04}
\bibliography{ijcai22}

\begin{thebibliography}{10}
\providecommand{\url}[1]{\texttt{#1}}
\providecommand{\urlprefix}{URL }
\providecommand{\doi}[1]{https://doi.org/#1}

\bibitem{bucak2011multi}
Bucak, S.S., Jin, R., Jain, A.K.: Multi-label learning with incomplete class
  assignments. In: CVPR 2011. pp. 2801--2808. IEEE (2011)

\bibitem{de2005tutorial}
De~Boer, P.T., Kroese, D.P., Mannor, S., Rubinstein, R.Y.: A tutorial on the
  cross-entropy method. Annals of operations research  \textbf{134}(1),  19--67
  (2005)

\bibitem{gao2015multi}
Gao, H., Nie, F., Li, X., Huang, H.: Multi-view subspace clustering. In:
  Proceedings of the IEEE international conference on computer vision. pp.
  4238--4246 (2015)

\bibitem{gretton2005measuring}
Gretton, A., Bousquet, O., Smola, A., Sch{\"o}lkopf, B.: Measuring statistical
  dependence with hilbert-schmidt norms. In: International conference on
  algorithmic learning theory. pp. 63--77. Springer (2005)

\bibitem{li2021concise}
Li, X., Chen, S.: A concise yet effective model for non-aligned incomplete
  multi-view and missing multi-label learning. IEEE Transactions on Pattern
  Analysis and Machine Intelligence  (2021)

\bibitem{lin2017focal}
Lin, T.Y., Goyal, P., Girshick, R., He, K., Doll{\'a}r, P.: Focal loss for
  dense object detection. In: Proceedings of the IEEE international conference
  on computer vision. pp. 2980--2988 (2017)

\bibitem{liu2013multi}
Liu, J., Wang, C., Gao, J., Han, J.: Multi-view clustering via joint
  nonnegative matrix factorization. In: Proceedings of the 2013 SIAM
  international conference on data mining. pp. 252--260. SIAM (2013)

\bibitem{tan2018incomplete}
Tan, Q., Yu, G., Domeniconi, C., Wang, J., Zhang, Z.: Incomplete multi-view
  weak-label learning. In: IJCAI. pp. 2703--2709 (2018)

\bibitem{tan2018multi}
Tan, Q., Yu, G., Domeniconi, C., Wang, J., Zhang, Z.: Multi-view weak-label
  learning based on matrix completion. In: Proceedings of the 2018 SIAM
  International Conference on Data Mining. pp. 450--458. SIAM (2018)

\bibitem{xu2013survey}
Xu, C., Tao, D., Xu, C.: A survey on multi-view learning. arXiv preprint
  arXiv:1304.5634  (2013)

\bibitem{zhang2013review}
Zhang, M.L., Zhou, Z.H.: A review on multi-label learning algorithms. IEEE
  transactions on knowledge and data engineering  \textbf{26}(8),  1819--1837
  (2013)

\bibitem{zhu2019improved}
Zhu, C., Miao, D., Zhou, R., Wei, L.: Improved multi-view multi-label learning
  with incomplete views and labels. In: 2019 International Conference on Data
  Mining Workshops (ICDMW). pp. 689--696. IEEE (2019)

\end{thebibliography}

\end{document}